\documentclass[letterpaper, 10 pt, conference]{ieeeconf}  

\IEEEoverridecommandlockouts                              

\usepackage{graphicx}
\usepackage{amsmath} 
\usepackage{amssymb}  
\usepackage{booktabs}
\usepackage{verbatim}
\usepackage{makecell}
\usepackage{amsmath}
\usepackage{algorithm}
\usepackage{algpseudocode}
\usepackage{amsmath}
\usepackage{caption}

\title{\LARGE \bf
Structured Pruning for Efficient Visual Place Recognition
}

\author{Oliver Grainge$^{1}$, Michael Milford$^{2}$, Indu Bodala$^{1}$, Sarvapali D. Ramchurn$^{1}$ and Shoaib Ehsan$^{1, 3}$%
\thanks{}%
\thanks{This work was supported by the UK Engineering and Physical Sciences Research Council through grants EP/Y009800/1 and EP/V00784X/1}
\thanks{$^{1}$O. Grainge, I. Bodala. S. D. Ramchurn and S. Ehsan are with the School of Electronics and Computer Science, University of Southampton, United Kingdon {\tt\small (email: oeg1n18@soton.ac.uk; i.p.bodala@soton.ac.uk; sdr1@soton.ac.uk; s.ehsan@soton.ac.uk).}}%
\thanks{$^{2}$M. Milford is with the School of Electrical Engineering and Computer Science, Queensland University of Technology, Brisbane, QLD 4000, Australia {\tt\small (email: michael.milford@qut.edu.au).}}%
\thanks{$^{3}$S. Ehsan is also with the School of Computer Science and Electronic Engineering, University of Essex, United Kingdom, {\tt\small (email: sehsan@essex.ac.uk).}}%
\thanks{Digital Object Identifier (DOI): see top of this page.}
}

\begin{document}

\maketitle
\thispagestyle{empty}
\pagestyle{empty}

\begin{abstract}

Visual Place Recognition (VPR) is fundamental for the global re-localization of robots and devices, enabling them to recognize previously visited locations based on visual inputs. This capability is crucial for maintaining accurate mapping and localization over large areas. Given that VPR methods need to operate in real-time on embedded systems, it is critical to optimize these systems for minimal resource consumption. While the most efficient VPR approaches employ standard convolutional backbones with fixed descriptor dimensions, these often lead to redundancy in the embedding space as well as in the network architecture. Our work introduces a novel structured pruning method, to not only streamline common VPR architectures but also to strategically remove redundancies within the feature embedding space. This dual focus significantly enhances the efficiency of the system, reducing both map and model memory requirements and decreasing feature extraction and retrieval latencies. Our approach has reduced memory usage and latency by 21\% and 16\%, respectively, across models, while minimally impacting recall@1 accuracy by less than 1\%. This significant improvement enhances real-time applications on edge devices with negligible accuracy loss. 
\end{abstract}

\section{Introduction}
Visual Place Recognition (VPR) is a critical capability in the field of robotics, enabling robots and devices to recognize locations they have previously visited based on visual inputs. This ability is fundamental to achieving global re-localization, which facilitates consistent mapping and localization across expansive areas within a visual perception system. Typically framed as an image retrieval problem, VPR requires a feature extraction model to generate embeddings that are proximate in feature space for images of the same location and distant for those of different locations. This is despite natural variations that occur in different images of the same place such as changes in lighting, occlusion, and appearance over time. 

For practical applications, such as in the loop closure module of Visual Simultaneous Localization and Mapping (VSLAM) systems, VPR methods must operate in real-time, necessitating the use of on-device processing \cite{quantizationvpr}. This requires efficient utilization of resources to minimize power and computational demands while maintaining accuracy and robustness across various environments. Recent methods, including TransVPR, AnyLoc, and DinoSalad, \cite{transvpr, anyloc, dinosalad} primarily focus on enhancing robustness and accuracy using large, powerful, self-supervised, pre-trained transformers. Although these methods achieve state-of-the-art accuracy, their resource demands are prohibitively high for real-time deployment on low powered edge devices \cite{quantizationvpr}.

\begin{figure}[t] 
    \centering
    \includegraphics[width=0.5\textwidth]{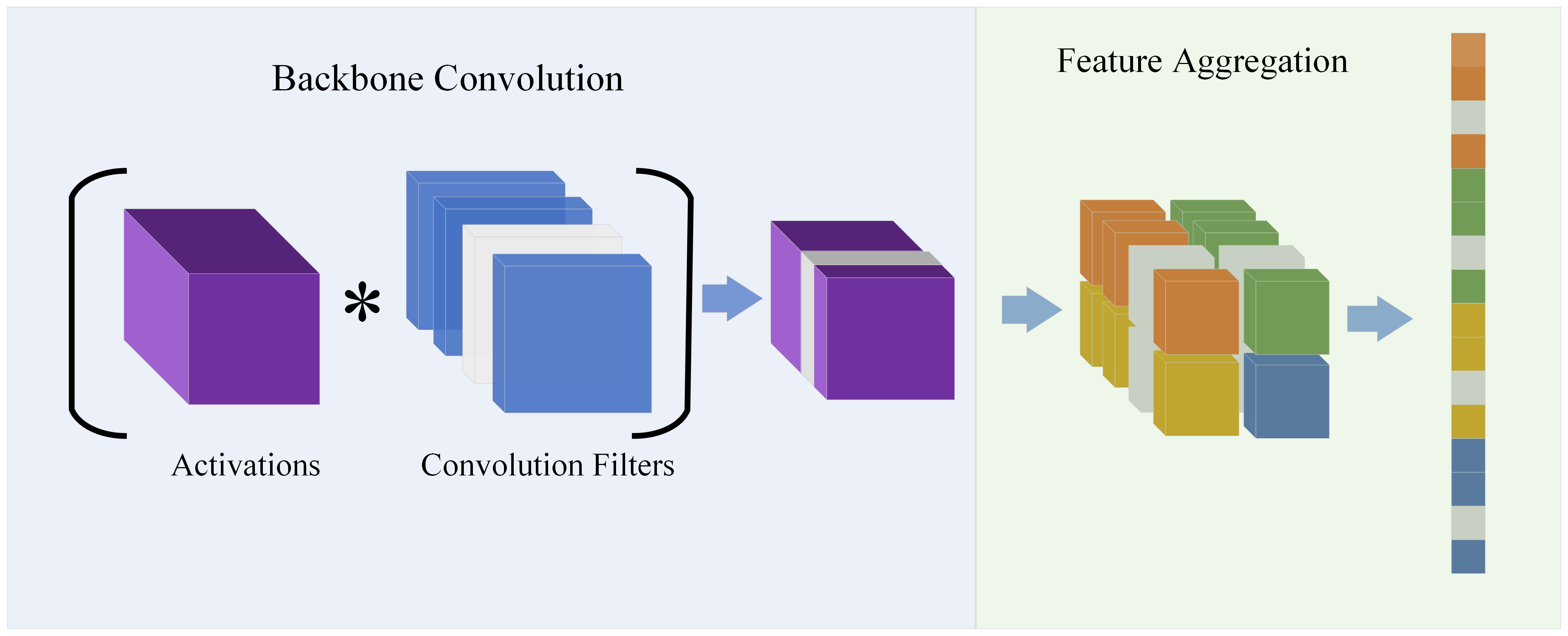}
    \caption{Structured Pruning of Convolution Visual Place Recognition Networks. In grey is the pruned backbone filters, which once removed simultaneously reduce the backbone size and descriptor dimension.}
    \label{fig:block}
\end{figure}

Current state-of-the-art VPR methods suited for real-time deployment, such as EigenPlaces, MixVPR, and CosPlace \cite{eigenplaces, mixvpr, cosplace}, utilize either ResNet or VGG convolutional backbones for feature extraction and employ fixed descriptor sizes. Although these methods achieve excellent recall scores and maintain efficient resource utilization, they do not fully exploit potential redundancies in their descriptors and feature extraction networks, which could be used to further enhance efficiency.

To address this gap and further exploit efficiency while maintaining robustness and accuracy, our work leverages structured pruning to optimize both the feature extraction networks and retrieval steps. Structured pruning involves removing groups of non-salient weights in the neural network that have minimal impact on accuracy \cite{pruning_survey}. This approach significantly reduces the network size, providing substantial resource savings. Unlike unstructured pruning, which removes individual weights, structured pruning eliminates entire neurons or filters in convolutional networks, resulting in a dense pruned model. This density preserves the efficiency of memory access and supports the acceleration of smaller networks through parallel computation \cite{pruning_acc}.

Structured pruning is most commonly applied to classification problems where the output dimension remains constant \cite{pruning_hardware}. However, in embedding models, it is both possible and computationally advantageous to reduce the output dimension. This reduction, however, can detrimentally impact the recall@1 score by decreasing the capacity of the embedding space. Therefore, identifying the minimal dimension of the embedding space and the optimal model architecture is critical to maximize recall@1 accuracy while minimizing memory and computational demands of VPR systems. Our method's approach to addressing this challenge is illustrated in Fig. \ref{fig:block}, where we use structured pruning both as an architecture search mechanism to identify an efficient model architecture and to minimize the embedding space. By executing these dual optimizations, our method provides a twofold performance benefit for the VPR system, reducing the resources needed for both feature extraction and map storage and searching. Furthermore, the architectures of current real-time VPR feature aggregation techniques, such as MixVPR, GeM, ConvAP, and NetVLAD are highly specialized \cite{mixvpr, gem, gsvcities, netvlad}. Thus, they require tailored approaches to maximize resource efficiency, and our methods are specifically designed to address this challenge.

Our key contributions include:

\begin{itemize}
    \item \textit{Introduction of Structured Pruning to VPR}: We introduce three new pruning methods to the VPR domain, adapting state of the art network architectures to eliminate non-essential weights.
    
    \item \textit{Enhanced Efficiency in Resource Utilization}: Our method achieves substantial reductions in resource demands, demonstrating an average reduction in memory and latency of 21\% and 16\% respectively for less than a 1\% decrease in recall@1.

    \item \textit{Comprehensive Assessment of Efficiency-Performance Trade-offs}: We analyze the trade-offs between efficiency and performance of our pruning methods, providing detailed results that include quantitative metrics and empirical embedding space analysis. Evaluations on an Nvidia Xavier NX platform, representative of typical robotic hardware, demonstrate our method's effectiveness.
    
\end{itemize}

This paper is organized as follows: Section II reviews related work on visual place recognition and pruning methods. Section III details our structured pruning approach for VPR architectures. Section IV presents the experimental setup and results, including analysis of memory usage, latency, and embedding space. Finally, Section V concludes with a discussion of our findings and their implications for real-time VPR systems.

\section{Related Work}
In this section, we review existing methods in visual place recognition and pruning, highlighting advancements and identifying gaps addressed by our approach.
\subsection{Visual Place Recognition}
To achieve high accuracy in VPR, methods must effectively extract features that are invariant to changes in different images of the same place. Convolutional Neural Networks (CNNs) excel in this area, benefiting from extensive research into optimal feature map pooling techniques, including hierarchical, spatial, and region of interest methods \cite{pyramid_pooling, cosplace, region_of_interest}. Beyond basic pooling, more advanced methods extract feature vectors and aggregate them into holistic representations. An example of which is the NetVLAD aggregation \cite{netvlad}, which uses a differential relaxation of the Vector of Locally Aggregated Descriptors (VLAD) technique to emphasize salient parts of the image. A more recent advancement is MixVPR \cite{mixvpr}, which adapts an MLP-Mixer architecture to iteratively process the channel and spatial dimensions of CNN feature maps.

Alongside these developments, the advent of powerful self-supervised foundation models has led to the emergence of highly robust VPR methods that adapt the general representations of the DINOv2 model to the VPR task \cite{dinov2}. Whilst all of them show excellent accuracy it remains a challenge to deploy them into the real-time perception systems of mobile robotic systems. Such systems usually default to using simple hand-crafted features \cite{orb} due to the need for resource efficiency in terms of memory consumption and processing latency. For example FAB-MAP \cite{fabmap}, ORB-SLAM2/3 \cite{orbslam3}, and Kimera1/2 \cite{kimera}, all use a bag of visual words aggregation despite the great improvements in accuracy recorded by deep methods. Furthermore as advancements in semantic scene understanding and dense mapping make their way into Spatial AI systems, the requirement for efficiency from every part of the system is further emphasized. In order to achieve it in the place recognition module, techniques such as Binarization, Quantization \cite{binaryvpr1, binaryvpr2} and Knowledge Distillation \cite{distillvpr, distillvpr2} have been explored. To the best of our knowledge however there has not been any work directly addressing the use of structured pruning in VPR and our work aims to address this gap.

\subsection{Pruning Methods}
It is well recognized that neural networks exhibit substantial redundancy in their weights \cite{imp}. Pruning capitalizes on this by removing weights that minimally affect task accuracy, thereby compressing the network and enhancing efficiency. This is achieved by reducing both the memory required to store the model weights as well as reducing the number of floating point operations (FLOPS) required in the forward pass. Pruning initially targeted individual weights \cite{obd}, resulting in models with high compression rates and minimal performance degradation. However, the irregularity of the resulting sparsity patterns from such approaches meant that without specialized accelerators the pruned models efficiency was not significantly different from it's dense counterpart \cite{structuredvsunstructured}. Only with specialized processors such as EIE \cite{eie} could these theoretical efficiencies be partially achieved. Pattern-based pruning addresses this limitation by imposing some structure on the pruned weights, such as requiring two out of every four consecutive weights to be zero, or enforcing blocks of weights to be pruned together \cite{nm_sparse}. These patterns can be exploited by specialized kernels and processing cores for acceleration. However the most general approach is structured pruning, which removes only groups of weights that would leave the resulting network a smaller dense version of itself. Examples include removing channels of linear layers or filters in convolutional layers \cite{pruning_survey, pruning_acc}. Pruning in this way enables effective memory saving and acceleration though general purpose parallel processors. 

Pruning can be performed on dense models before, during, or after training \cite{pruning_survey}. It is most commonly applied in a one-shot manner, involving a single round of post-training pruning followed by fine-tuning. While this method is simple and effective, higher compression rates can be achieved through iterative rounds of pruning and fine-tuning \cite{imp}. Although it was previously considered less efficient to train models in this manner, recent evidence has shown that by selecting the right learning schedule, training timescales can be equivalent \cite{lr_schedule}.

While network pruning has been applied to numerous tasks in computer vision, including classification, object detection, segmentation, depth estimation, and more, a common requirement across these applications is that the output dimension must remain the same post-pruning \cite{pruning_survey}. However, this is not a requirement in VPR, where the embedding dimension can vary. Consequently, pruning can be strategically used to optimize both feature extraction and retrieval stages. While other techniques such as principle components analysis (PCA) have been used to reduce the dimensionality of the descriptor \cite{netvlad}, they do not jointly optimize feature extraction and retrieval efficiency. Therefore, our work is crucial for fully optimizing all sources of resource consumption in VPR systems and thereby enabling their deployment in real-time perception systems on low-powered and low-cost commodity hardware.

\section{Pruning VPR Architectures}\label{sec:pruning}
Despite the broader representational capacity of transformers, convolutional architectures are preferred in lower-powered embedded systems due to their efficiency in latency and memory \cite{quantizationvpr, geobenchmark, transformer_compression}. CNNs achieve computational efficiency through weight sharing, hierarchical processing, and require less data to train due to their inductive biases, making iterative pruning more resource-efficient. In subsections \ref{sec:fully_conv}, \ref{sec:mixvpr} and \ref{sec:netvlad} we present our three novel methods for pruning state-of-the-art convolutional architectures for VPR.

\subsection{Fully Convolutional Models}\label{sec:fully_conv}
\begin{algorithm}
\caption{Pruning Fully Convolutional Modelss}\label{alg:fully_conv}
\begin{algorithmic}[1]
\State \textbf{Input:} Convolutional weights $W^l$, intermediate pruning rate $r_{i}$, output layer pruning rate $r_{o}$

\For{$l = 1$ to $L-1$} \Comment{Pruning intermediate layers}
    \State Compute $\mathbf{i}$: indices of the smallest $r_{i} \times c_o \text{ filters of } W^l$ based on $\ell_1$ norm
    \State $W^l \leftarrow W^l_{c_o \notin \mathbf{i},:,:,:}$ \Comment{Remove smallest output filters}
    \State $W^{l+1} \leftarrow W^{l+1}_{:, c_i \notin \mathbf{i}, :, :}$ \Comment{Remove corresponding input filters in the next layer}
\EndFor

\State Compute $\mathbf{i}$: indices of the smallest $r_{o} \times c^L_0\text{ filters in } W^L$ based on $\ell_1$ norm
\State $W^L \leftarrow W^L_{c_o \notin \mathbf{i},:,:,:}$ \Comment{Remove smallest output filters in the last layer of the CNN backbone}

\State \textbf{Output:} Pruned fully convolutional model.
\end{algorithmic}
\end{algorithm}

We begin by considering fully convolutional networks, as they are the most lightweight. This category includes the CosPlaces \cite{cosplace}, EigenPlaces \cite{eigenplaces}, and ConvAP \cite{gsvcities} architectures. All three use a ResNet50 backbone, with EigenPlaces and CosPlaces incorporating GeM \cite{gem}, and ConvAP utilizing spatial average pooling \cite{gsvcities}. To effectively prune channels in these architectures, the approach outlined in Algorithm \ref{alg:fully_conv} can be used. Here, \( c_i \) and \( c_o \) refer to the input and output channels of the convolution, respectively, and \( W^L \) represents the last convolutional weight in a network with \( 1 \rightarrow L \) backbone layers. 

For pruning the aggregation module, such as GeM \cite{gem}, as implemented in EigenPlaces and CosPlace \cite{eigenplaces, cosplace}, no changes are required. GeM pooling, as shown in Equation \ref{eqn:gem}, has a single learnable scalar \( p \), which is independent of the backbone's output feature map dimension. This allows the final pruning rate of \( W^L \) (\( r_o \)) to control the output feature dimension.

\begin{equation}\label{eqn:gem}
    y_c = (\frac{1}{H W} \sum^{H}_{h=0} \sum^{W}_{w=0} X^L_{c,h,w})^{\frac{1}{p}}
\end{equation}

Spatial Average Pooling, as utilized in the ConvAP VPR model, is described by Equation \ref{eqn:convap}. In this method, \( P \) represents the spatial pooling block size, and \( X^L \) denotes the last convolution activation. Being a parameter-less pooling method, it is also independent, and its descriptor dimension can likewise be controlled by the output pruning rate $r_o$.

\begin{equation}\label{eqn:convap}
    y_{j,k \cdot c_o} = \frac{1}{P^2}\sum^{P-1}_{m=0} \sum^{P-1}_{n=0} X^{L}_{c_o,i*P+m, j*P+n}
\end{equation}

\subsection{MixVPR}\label{sec:mixvpr}
\begin{algorithm}
\caption{Pruning the MixVPR Module}\label{alg:mixvpr}
\begin{algorithmic}[1]
\State \textbf{Input:} Convolutional features $X^L$, weights $W_1$, $W_2$, $W_d$, $W_r$
\State \textbf{Output:} Pruned MixVPR module
\State Flatten the spatial dimensions: $F = X^L \rightarrow \mathbb{R}^{c_0, hw}$
\State Process feature vectors through $N$ MLP blocks:
\begin{equation}\label{eqn:mlpblock}
    F^{n+1} = W_2 \sigma(W_1 F^n) + F^n
\end{equation}
\State Aggregate features:
\begin{equation}\label{eqn:mixerblock_depth}
    F_d = W_{d}(\text{Transpose}(F^N))
\end{equation}
\begin{equation}\label{eqn:mixerblock_row}
    y = \text{flatten}(W_{r}(\text{Transpose}(F^N)))
\end{equation}
\State Prune $W_{d}\left[:, :h \notin G\right]$ \text{ where } $G$ \text{ is the pruned output filters of } $W^L$
\State Compute $\mathbf{i}$: indices of the smallest $r_{o} \times \text{number of channels in } W_d$ based on $\ell_1$ norm
\State $W_d \leftarrow W_{d}\left[c \notin \mathbf{i}, :\right]$ \Comment{Prune channels}
\State \textbf{Output:} Pruned MixVPR model.
\end{algorithmic}
\end{algorithm}

MixVPR integrates an MLP-Mixer aggregation that first flattens the spatial dimensions of convolutional features as described in step 3 of Algorithm \ref{alg:mixvpr}. This operation produces feature vectors $F$, processed through $N$ residual MLP blocks (Equation \ref{eqn:mlpblock}), where $\sigma$ denotes the ReLU activation function. This enables each feature vector’s linear projection to incorporate context from the entire input image. Our structural pruning method for $W^L$, the last convolutional layer (Algorithm \ref{alg:fully_conv}), reduces the number of feature vectors in $F$ without changing their dimensionality $hw$, leaving $W_1$ and $W_2$ unaffected. Consequently, the MLP block stage’s computational cost decreases linearly with the pruned output channels $c^L_o$, greatly improving the networks efficiency.

To aggregate the features $F^N$, MixVPR applies Equations \ref{eqn:mixerblock_depth} and \ref{eqn:mixerblock_row} sequentially, creating a dependency between $W_d$ and $W^L$ due to the transposition operation, necessitating their joint pruning. Additionally, to reduce the final descriptor dimension, we strategically prune the output channels of $W_d$. However, to maintain the integrity of MixVPR’s four-dimensional row projection and prevent performance losses, our method avoids pruning $W_r$. This ensures that efficiency gains do not compromise the model's capacity.

\subsection{NetVLAD}\label{sec:netvlad}
\begin{algorithm}
\caption{Pruning the NetVLAD Module}\label{alg:netvlad}
\begin{algorithmic}[1]
\State \textbf{Input:} Feature vectors $T_i$, cluster centers $C_z$, pruning rate $r_o$
\State \textbf{Output:} Pruned NetVLAD model
\State Extract and aggregate feature vectors:
\begin{equation}\label{eqn:netvlad}
    V_z = \sum^N_{i=1} a_{i,z}(T_i - C_z)
\end{equation}

\State Apply K-means clustering on \ensuremath{C} to prune cluster centers.
\State Set the new cluster number to pruning rate \ensuremath{r_o \times} the number of clusters in \ensuremath{C}.
\State \textbf{Output:} Pruned NetVLAD model.
\end{algorithmic}
\end{algorithm}

Unlike MixVPR, NetVLAD extracts convolutional feature vectors \( T_i \) across the channel dimension \( \mathbb{R}^{HW, c_o} \) and aggregates them by summing residuals between \( T_i \) and softly assigned cluster centers \( C_z \). Equation \ref{eqn:netvlad} shows this process, where \( V_z \) is the aggregated descriptor for the z-th cluster, and \( a_{iz} \) is the soft assignment weight of \( T_i \) to cluster \( C_z \).

The NetVLAD descriptor dimension is the product of the feature vector size \( T_i \) and the number of clusters \( C \) (128 dimensions with 64 clusters originally \cite{netvlad}). To reduce memory consumption and computation, we prune the number of clusters, adjusting the channel dimension of \( C \) as shown in Algorithm \ref{alg:netvlad}. Instead of using group norm importance for pruning, we apply K-means clustering to reduce cluster centers based on the pruning rate. This method preserves cluster distribution and reduces the descriptor dimension linearly, crucial for managing NetVLAD's large 8192-dimensional descriptors.

\section{Experimental Setup}

This section outlines the experimental framework used to evaluate the structured pruning approach for VPR within the real-time operational constraints typical of lightweight mobile robots. The experiments are conducted using the Nvidia Xavier NX embedded system, selected to represent the computational limitations and capabilities of modern autonomous systems, such as drones. The following subsections describe the specific methodologies for structured pruning, training regimes, and the metrics used to evaluate resource and efficiency enhancements in VPR systems.

\subsubsection{Structured Pruning}
Our method employs iterative magnitude pruning (IMP) on a pre-trained VPR model, using the algorithms outlined in Section \ref{sec:pruning}. We choose iterative magnitude pruning \cite{imp} over one-shot pruning because it impacts the network gradually, thereby enhancing error recovery and improving generalization \cite{imp}.

\begin{figure}[t] 
    \centering
    \includegraphics[width=0.5\textwidth]{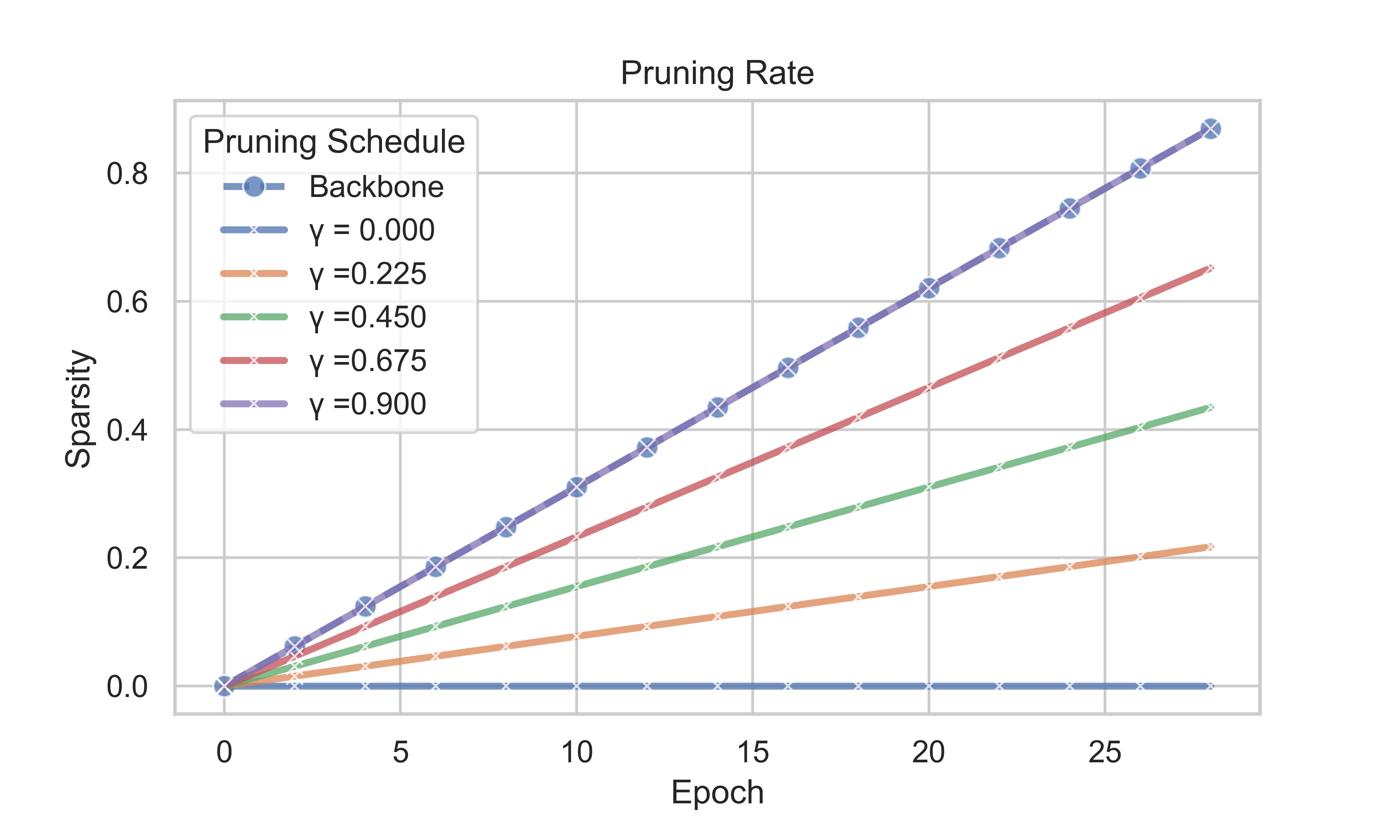}
    \caption{Linear Pruning Schedule Overview. This schedule shows the backbone pruning schedule ending with a final sparisty of 0.9. The aggregation pruning hyper-parameter $\gamma$ represents the final aggregation and descriptor sparsity, regulating the sparsity ratio between the network's backbone and the descriptor throughout each step of the pruning process.}
    \label{fig:pruning}
\end{figure}

Since our method prunes both the aggregation method and backbone width, we introduce a hyperparameter $\gamma$ to balance the pruning ratios $r_i$ and $r_o$ between the network's backbone and aggregation module. This parameter is crucial as it modulates the trade-off between efficiency gains in feature extraction and retrieval. Pruning the aggregation module reduces the descriptor dimension, decreasing both map memory usage and retrieval latency, while backbone pruning reduces model memory and the latency consumed by feature extraction. The hyperparameter $\gamma$ controls this trade-off by determining the final sparsity value in the linear pruning rate schedule. The effect is illustrated in Fig. \ref{fig:pruning}, where a $\gamma$ of $0.9$ leads to the same sparsity in the descriptor and backbone, whereas a $\gamma$ of $0.0$ prevents any pruning of the descriptor.

\subsubsection{Training}
We conduct all our training experiments using a single Nvidia A6000 GPU. Each model undergoes training on the GSV-Cities dataset, employing multi-similarity loss with a batch size of 120 over 30 epochs \cite{gsvcities}. We utilize the Adam optimizer, initiating at a learning rate of $1\times10^{-3}$ and applying a multiplier of $0.3$ every 5 epochs to adjust the rate.

After training the NetVLAD, ConvAP, MixVPR and GeM models \cite{netvlad, gsvcities, mixvpr, eigenplaces} we begin our IMP fine-tuning run using our pruning methods outlined in Section \ref{sec:pruning}. We prune all prune-able layers in the backbone to 90\% sparsity pruning every 2 epochs for a total of 50 epochs in total. We also prune the aggregation methods using our pruning techniques outlined in Section \ref{sec:pruning} to $(\gamma * 100) \%$ according to the pruning schedules outlined in Fig. \ref{fig:pruning}. Again similar to the initial training, we use a learning rate decay multiplier with a value of $0.3$ but initialize the learning rate at $1\times10^{-4}$ for fine-tuning and reset the schedule after each pruning round. 

\subsection{Memory}
For embedded systems memory consumption is a critical concern. In order, to maintain VPR system latency constraints, the model and map must be kept in unified system memory (DRAM) which for a powerful embedded system such as the Nvidia Xavier Nx is 8Gb and for other devices may be much smaller.

\begin{figure}[t] 
    \centering
    \includegraphics[width=0.5\textwidth]{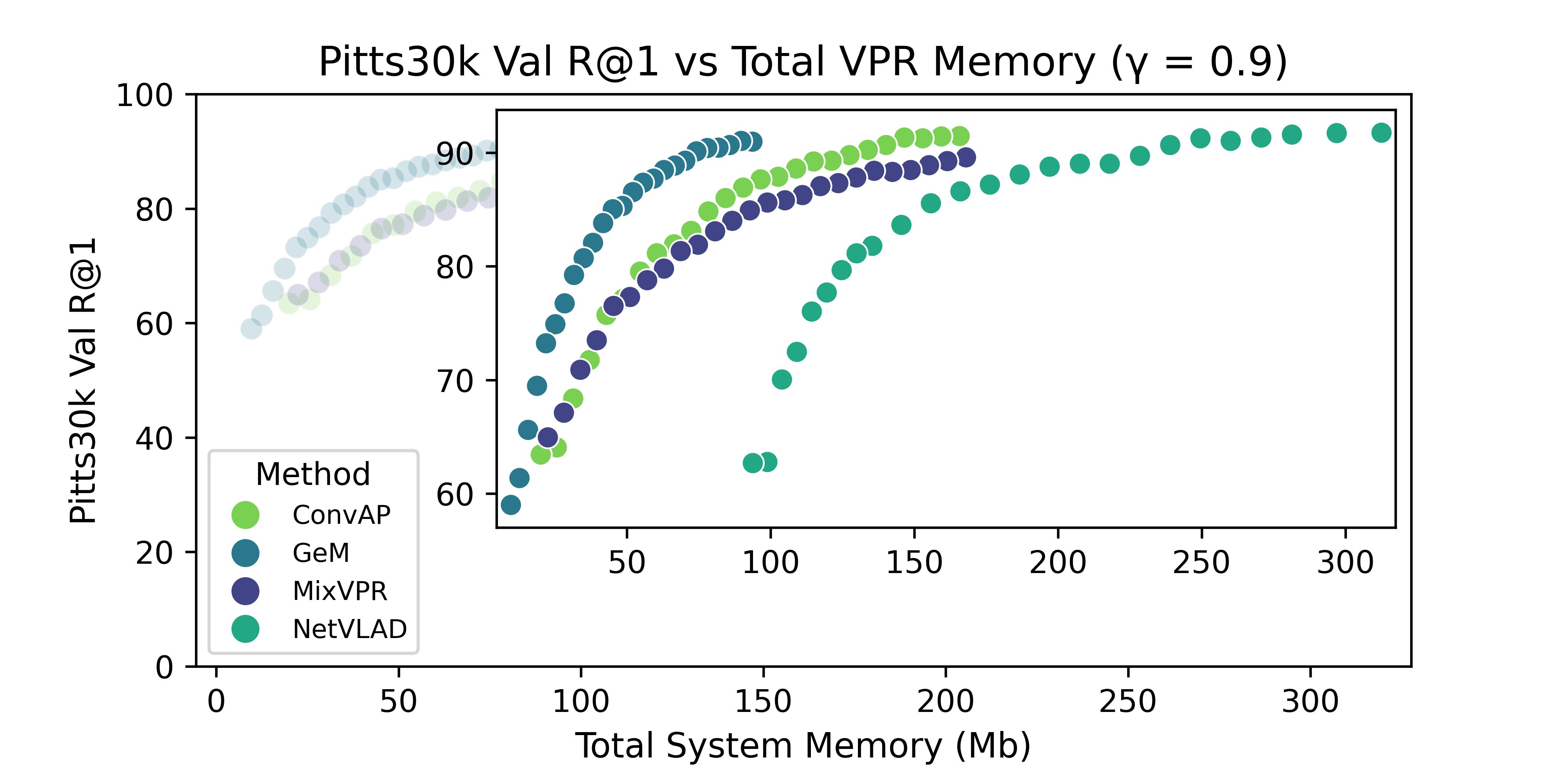}
    \caption{Total memory of the VPR system including the sum of the model and map embedding consumptions against the recall@1 score. The curves are created by iterative magnitude pruning of the feature extraction network.  }
    \label{fig:memory}
\end{figure}

As seen in Fig. \ref{fig:memory} there is a performance trade-off for all architectures between recall@1 accuracy and memory consumption. Each model shows that under small levels of pruning the recall@1 accuracy drop is minimal in comparison to memory reduction, thereby showing pruning's ability to produce memory savings. This is particularly the case for NetVLAD, showing just a 0.7\% recall@1 score drop for a 58Mb drop in memory consumption, which is likely due to the redundancy present in its large descriptor size and cluster centers. This is similarly the case for ConvAP, GeM and MixVPR which show a 14\%, 9\% and 12\% reduction in memory consumption for less than a 1\% drop in recall@1 produced via IMP. Interestingly ConvAP has a higher initial recall@1 score than MixVPR despite requiring a similar amount of memory. During pruning however the performance of ConvAP drops off quicker showing that the pruned architecture has a slightly reduced memory efficiency trade-off. 

\subsection{Latency}

\begin{figure}[h]
    \centering
    \includegraphics[width=0.5\textwidth]{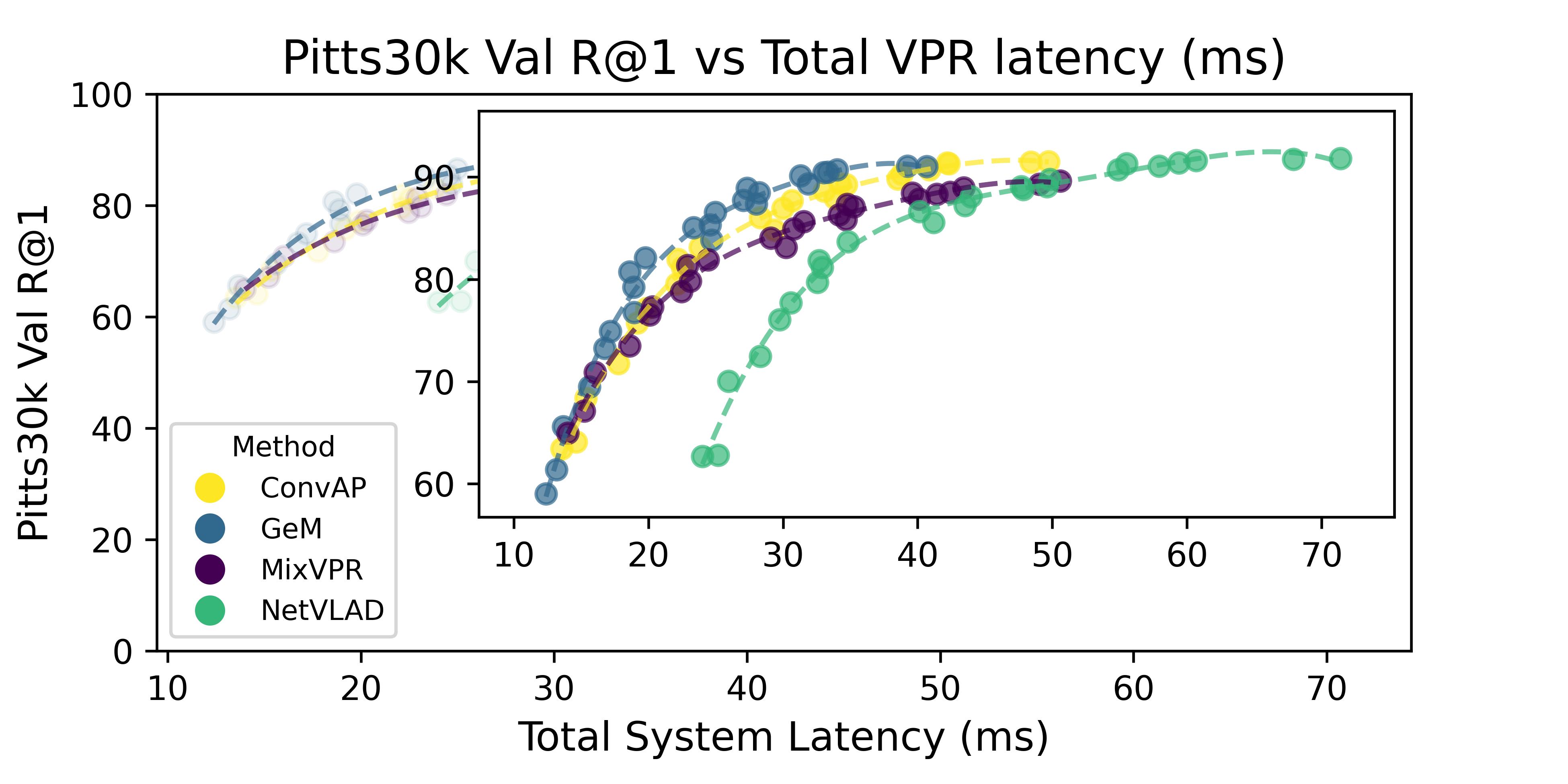}
    \caption{Total latency of the VPR system including the feature extraction and matching latencyies of a single image against the recall@1 score. The curves are created by iterative magnitude pruning of the feature extraction network.}
    \label{fig:latency}
\end{figure}

Optimizing latency is crucial for VPR systems because high latency can result in missed loop closures, increased drift error, or processing bottlenecks \cite{quantizationvpr}. Lower VPR inference latency also allows for deployment on cost-effective hardware with less memory and bandwidth or processing parallelism. Given that visual place recognition systems operate at up to 20Hz \cite{kimera}, achieving this rate is essential.

Figure \ref{fig:latency} demonstrates that pruning can significantly reduce the total system latency of the VPR method below the real-time perception constraint of 20Hz (50ms). The latency result displayed in this figure is measured for the feature extraction and matching of a single keyframe from the Pittsburgh 30K validation dataset with GPU acceleration. Fig. \ref{fig:latency} also reveals that the recall@1 and system latency exhibit a trade-off curve similar to the memory efficiency shown in Fig. \ref{fig:memory}, suggesting a positive correlation between memory and latency metrics under our pruning methods, while emphasizing minimal trade-offs between them. Regarding latency measurements, at high levels of pruning, it can be significantly lowered to 30-45ms, however the recall@1-latency gradient is steeper. In contrast, at lower pruning rates, the curve remains relatively flat, indicating that structural pruning across all network architectures can enhance VPR system latency with minimal impact on R@1 accuracy, but it is more effective at lower pruning ratios. For instance, during pruning, the NetVLAD architecture can reduce system latency from 70ms to 50ms with only a 0.3\% reduction in recall@1, effectively meeting the required latency constraint.

\subsection{Aggregation Pruning Rate}
To balance the trade-off between joint optimization of the feature extraction resource consumption and that of the retrieval step, we perform experimentation with 5 different $\gamma$ rates, controlling the trade-off between retrieval and extraction efficiencies. 

\begin{figure}[h]
    \centering
    \includegraphics[width=0.5\textwidth]{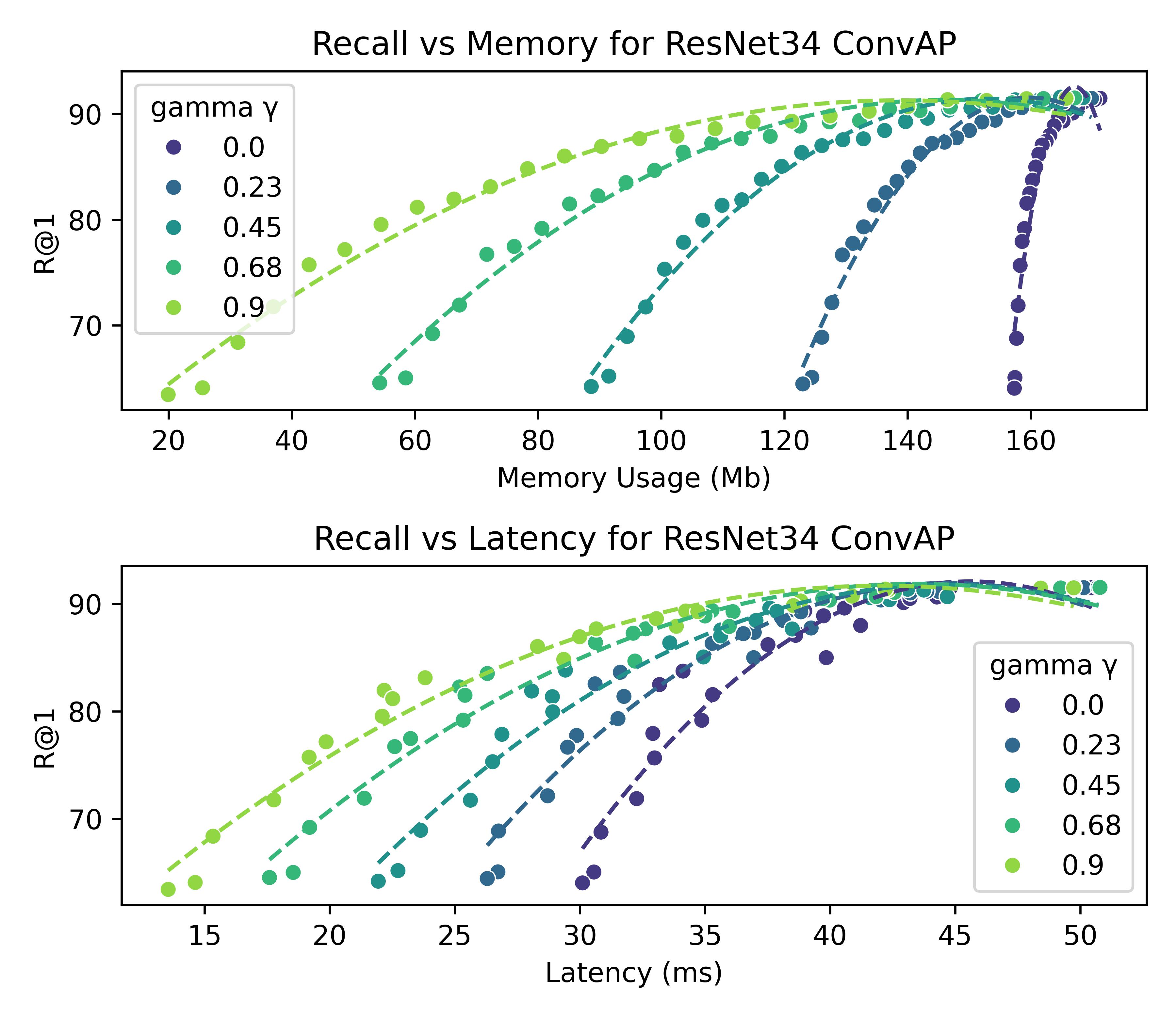}
    \caption{Efficiency-recall@1 trade-off curves for the pruned ConvAP VPR method \cite{gsvcities} on the on Pitts30k Validation dataset. The hyper-parameter $\gamma$ shows the effect of altering the pruning ratio between the backbone and feature aggregation.}
    \label{fig:pruning_ratio}
\end{figure}

Figure \ref{fig:pruning_ratio} shows the effect of $\gamma$ on the trade-off between recall@1, memory and latency during pruning. The bottom graph shows a strong trend with higher levels of descriptor pruning improving the recall@1-latency trade-off. This is because as seen by Fig. \ref{fig:bar} the retrieval step provides a substantial source of latency for the VPR system, and pruning of the aggregation module can effectively reduce this without having a large impact on the recall@1 score, thereby emphasizing the redundancy present in the embedding space. The top graph of Fig. \ref{fig:pruning_ratio} shows a similar trend with an even more significant impact. Again, higher levels of pruning in the descriptor lead to an improved memory efficiency trade-off. This is due to the significant memory consumption caused by storing the map of descriptors. This highlights the importance of using pruning to minimize the descriptor dimension. 

\section{Resource Consumption Analysis}
The total resource consumption of VPR systems comes from both the feature extraction and matching sub-tasks. In order to optimize the resource utilization of the system as a whole it is important to understand how much each sub-task contributes to resource usage and which type of resource usage.

\begin{figure}[h]
    \centering
    \includegraphics[width=0.5\textwidth]{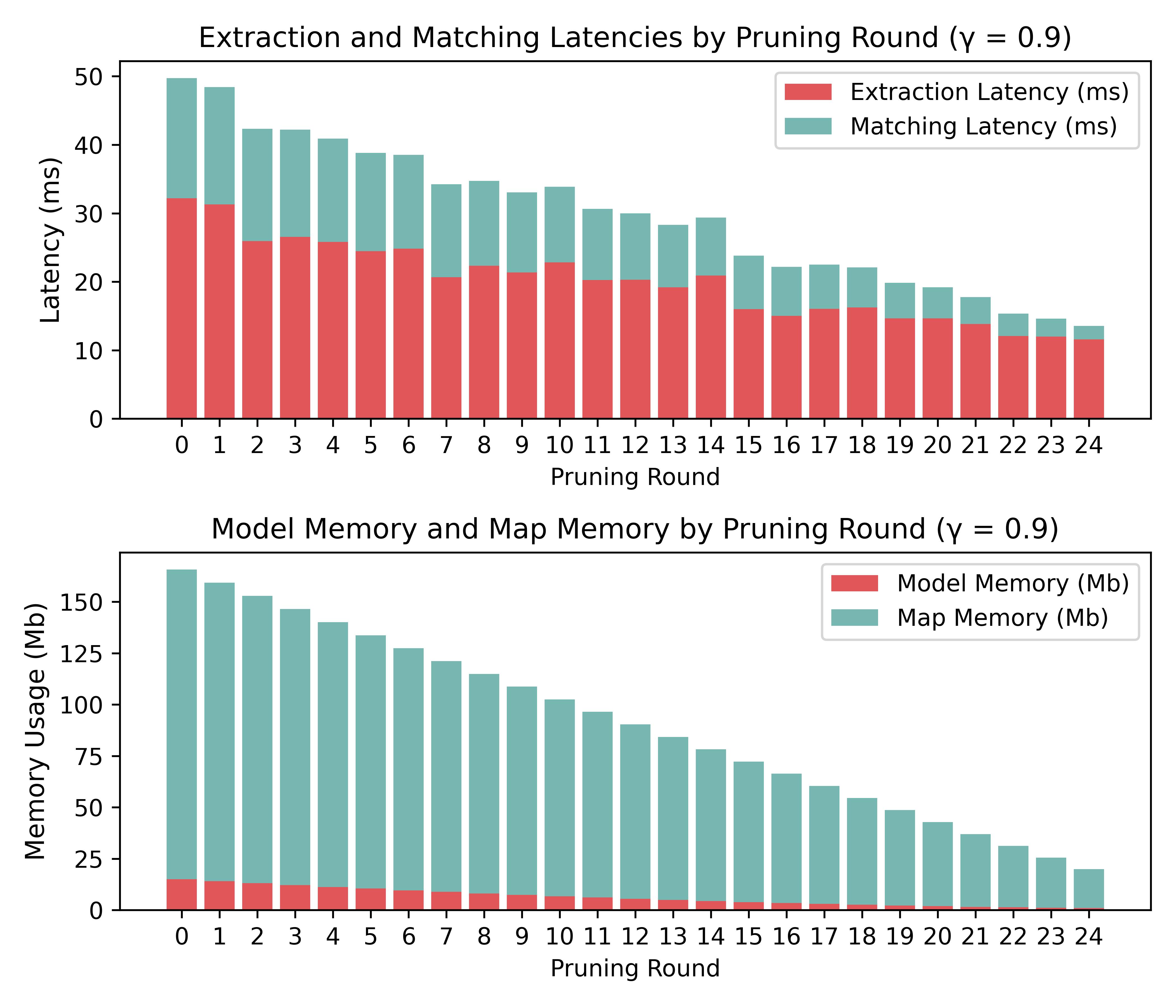}
    \caption{Breakdown of the Memory and Latency consumption for the ConvAP \cite{gsvcities} VPR system measured across pruning rounds: this table illustrates the impact of pruning on the resource consumption of feature extraction and matching sub-tasks measured on the Pitts30k Validation Dataset.}
    \label{fig:bar}
\end{figure}

From Fig. \ref{fig:bar} it can be seen that IMP is an effective way of reducing both latency and memory consumption. However the contributions from each part of the network are different. For instance, in the lower bar chart, it is seen that the main source of the VPR system memory consumption is in the map storage, whereas for latency it is is the model extraction. Therefore should memory be more of a concern for the target device, descriptor pruning will provide a more significant improvement, whereas for throughput and bandwidth limited devices where latency is high, backbone pruning will be more advantageous.

\begin{table*}[t]
\caption{Resource consumption and latency of VPR architectures at 40\% sparsity. The best metrics per column are highlighted in bold, demonstrating the trade-off between accuracy and resource efficiency.}
\label{tab:sparsity_data}
\centering
\begin{tabular}{|l|c|c|c|c|c|c|c|c|c|}
\toprule
\hline
Method & $\gamma$ & \makecell{Extraction \\ Latency (ms)} & \makecell{Matching \\ Latency (ms)} & \makecell{Map Memory \\ (Mb, 10,000 \\ embeddings)} & \makecell{Model \\ Memory (Mb)} & \makecell{Pitts30K \\ R@1} & \makecell{Pitts250K \\ R@1} & \makecell{MSLS \\ R@1} & \makecell{SpedTest \\ R@1} \\
\midrule
\hline
\hline
ConvAP & 0.00 & 25.0 & 18.2 & 156.2 & 9.7 & 90.5 & 89.8 & 73.1 & 79.6 \\
ConvAP & 0.23 & 25.0 & 17.0 & 146.6 & 9.7 & 90.4 & 89.8 & 73.1 & 79.6 \\
ConvAP & 0.45 & 25.5 & 16.9 & 137.0 & 9.7 & 90.4 & 89.6 & 73.5 & 80.4 \\
ConvAP & 0.68 & 24.9 & 14.8 & 127.4 & 9.7 & 90.5 & 89.4 & 73.2 & 79.9 \\
ConvAP & 0.90 & 24.8 & 13.7 & 117.8 & 9.6 & 89.8 & 89.5 & 73.2 & 79.7 \\
\hline
GeM & 0.00 & 21.1 & 8.9 & 78.1 & 11.6 & 89.1 & 88.8 & 73.0 & \textbf{82.0} \\
GeM & 0.23 & \textbf{21.0} & 8.2 & 72.6 & 11.4 & 89.0 & 88.5 & 73.2 & 81.7 \\
GeM & 0.45 & 25.0 & 7.7 & 68.5 & 12.0 & 89.5 & 89.3 & 74.5 & 81.2 \\
GeM & 0.68 & 25.0 & 7.2 & 63.7 & 11.8 & 89.3 & 89.3 & 74.7 & 80.2 \\
GeM & 0.90 & 25.2 & \textbf{6.7} & \textbf{58.6} & 11.6 & 89.3 & 89.1 & 74.6 & 81.1 \\
\hline
MixVPR & 0.00 & 21.8 & 19.7 & 156.2 & 11.2 & 87.7 & 85.6 & 78.8 & 67.7 \\
MixVPR & 0.23 & 22.1 & 17.4 & 146.5 & 11.2 & 87.6 & 85.6 & 78.6 & 68.5 \\
MixVPR & 0.45 & 21.9 & 15.8 & 136.7 & 11.2 & 87.4 & 85.5 & 77.8 & 68.7 \\
MixVPR & 0.68 & 22.1 & 14.7 & 127.0 & 11.2 & 87.4 & 85.3 & 77.8 & 67.5 \\
MixVPR & 0.90 & 21.8 & 13.0 & 112.3 & 11.2 & 87.3 & 85.3 & 77.8 & 67.9 \\
\hline
NetVLAD & 0.00 & 29.0 & 32.1 & 278.3 & 9.4 & \textbf{91.0} & \textbf{90.3} & 81.8 & 73.8 \\
NetVLAD & 0.23 & 29.3 & 32.8 & 278.3 & 9.4 & 91.0 & 90.2 & \textbf{82.7} & 73.5 \\
NetVLAD & 0.45 & 28.5 & 29.3 & 253.9 & 9.4 & 90.8 & 89.9 & 82.2 & 74.6 \\
NetVLAD & 0.68 & 28.6 & 28.1 & 244.1 & 9.4 & 90.7 & 89.8 & 82.0 & 75.9 \\
NetVLAD & 0.90 & 28.4 & 26.5 & 229.5 & \textbf{9.4} & 90.7 & 89.5 & 81.6 & 75.0 \\
\hline
\bottomrule
\end{tabular}\label{tab:results}
\end{table*}

\subsection{Architecture Comparison}
Table \ref{tab:results} presents the resource consumption and accuracy results on the Pittsburgh 30k and 250k datasets \cite{netvlad}, the Mapillary Street-Level Sequences (MSLS) dataset \cite{msls}, and the SpedTest dataset \cite{sped}. The results are computed at a sparsity level of 40\%. When comparing the architectures and their respective pruning methods, NetVLAD consistently performs the best, achieving the highest recall@1 scores on three of the four benchmark datasets. This performance is likely due to the significant redundancy in its cluster centers after the pre-training stage, which enhances its robustness under pruning. However, NetVLAD's descriptor is the largest, resulting in higher map memory and matching latency resource consumption. Additionally, although its model memory consumption is low, the aggregation method, which requires summing the residuals assigned to clusters, leads to the highest extraction latency.

GeM, the smallest model in terms of descriptor size, surpasses MixVPR in performance under our pruning methods, particularly on the SpedTest dataset. GeM's ability to generalize to other test distributions outside the training distribution, such as SpedTest, is significantly better than that of MixVPR. The fully convolutional model ConvAP shows similar generalization, suggesting that the MLP layers in MixVPR might be overfitting to the training distribution. Overall, on average with our methods the models experience less than a 1.2\% performance degradation at moderate pruning levels (40\%) on the Pitts30k benchmark, thereby demonstrating the effectiveness of our methods in removing redundancy from the models.

\subsection{Embedding Space Procrustes Analysis}

\begin{figure}[h]
    \centering
    \includegraphics[width=0.5\textwidth]{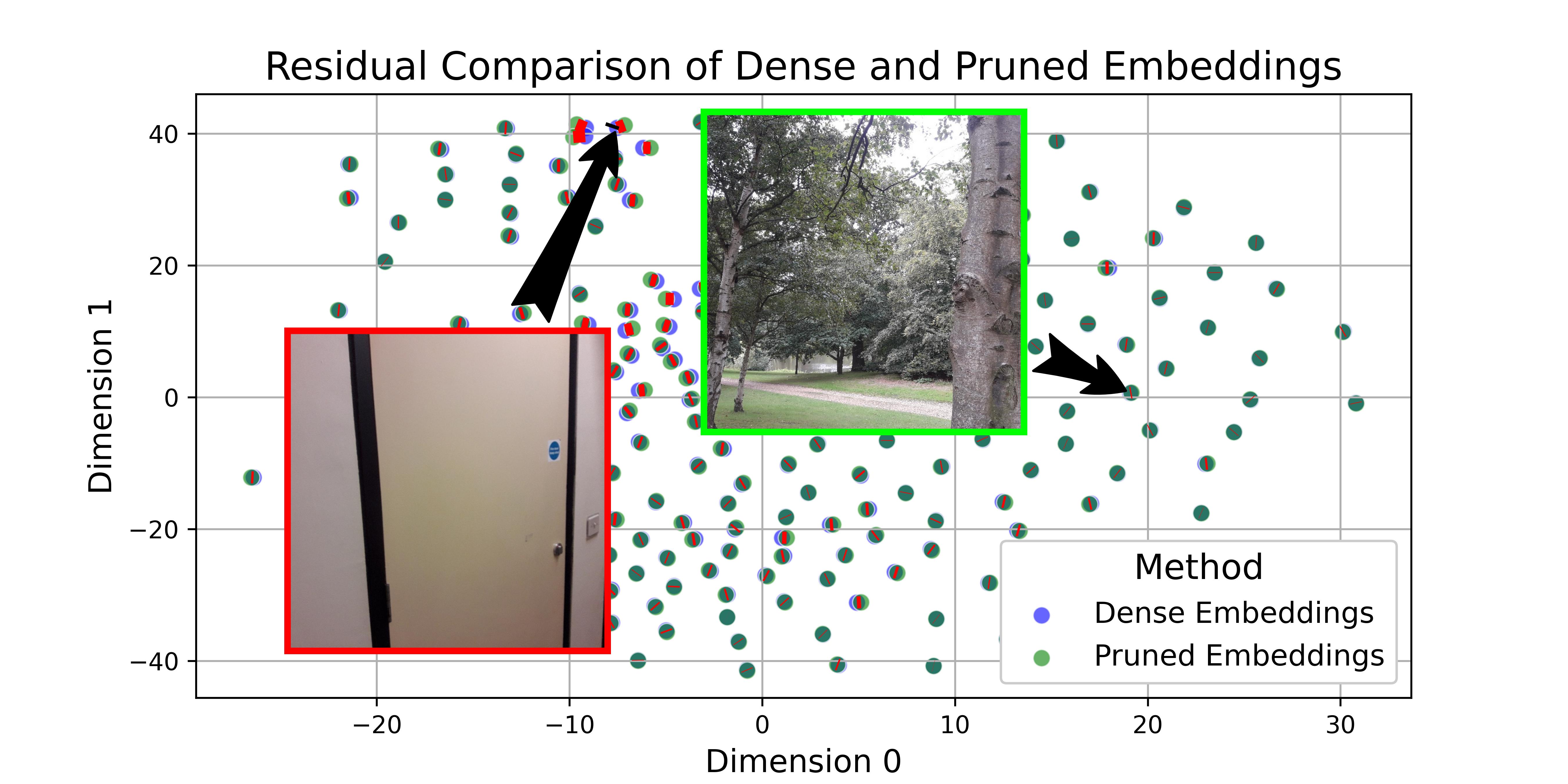}
    \caption{Embedding space of both the dense and pruned VPR models. Shown in red is the residuals between the same image embedded with different models. The image bordered in red has the greatest embedding space residual and in green, the smallest. The figure shows pruning changes the embedding space most for images with perceptual aliasing.}
    \label{fig:embedding}
\end{figure}

While the recall@1 score is a standard metric for VPR accuracy, it’s crucial to consider other factors, especially in deployment environments. A system that performs poorly in less common environments can significantly reduce its practical usability. To address this, we empirically investigate how our pruning methods affect the embedding space and identify the environments where they might fail.

We begin by selecting a ConvAP model pruned to 40\% with a $\gamma$ of 0.9 and compare it to its dense counterpart using embedding space analysis. The dense model's large descriptor dimension is aligned with the pruned model using PCA linear projection, followed by Procrustes analysis \cite{procrustes} for further refinement. These aligned spaces are then projected into two dimensions using t-SNE \cite{tsne}, as shown in Fig. \ref{fig:embedding}. The figure reveals substantial overlap between the dense and pruned model embeddings, indicating that pruning has minimal impact on the overall metric space. To explore subtle differences, we highlight residuals between the same image embeddings in red. Further analysis focuses on images with the largest and smallest residuals, marked in red and green, respectively. Fig. \ref{fig:embedding} shows that images with high perceptual aliasing have the largest residuals, suggesting pruned models may struggle in environments with repetitive structures like indoor scenes. However, in more distinct environments such as woodlands, the pruned models perform robustly

\section{CONCLUSION}
This paper presents a novel method for structured pruning of VPR systems, aiming at significantly enhancing efficiency without sacrificing accuracy. By employing our architecture-specific pruning methods, we demonstrated that we can on average across all models effectively reduce memory usage and latency by 21\% and 16\%, respectively, with a marginal decrease in recall@1 accuracy of less than 1\%. Additionally, our findings indicate that the highest level of redundancy exists within the embedding space compared to the model itself, suggesting that the aggregation module should be pruned more aggressively. These results underscore the feasibility of deploying efficient VPR systems on low-cost resource-constrained platforms, such as those used by light-weight mobile robots. The structured pruning technique introduced here ensures that the crucial capabilities of robust recognition are maintained, thus paving the way for practical, real-time applications of VPR technologies.

\bibliographystyle{IEEEtran}
\bibliography{references}

\end{document}